\documentclass[conference]{IEEEtran}

\IEEEoverridecommandlockouts
\usepackage{amsmath,amssymb,amsfonts}
\usepackage{mathtools}
\usepackage{algorithmic}
\usepackage{graphicx}
\usepackage{textcomp}
\usepackage{xcolor}
\usepackage{array}
\newcolumntype{P}[1]{>{\centering\arraybackslash}p{#1}}
\newcolumntype{M}[1]{>{\centering\arraybackslash}m{#1}}
\def\BibTeX{{\rm B\kern-.05em{\sc i\kern-.025em b}\kern-.08em
    T\kern-.1667em\lower.7ex\hbox{E}\kern-.125emX}}
\begin{document}

\title{ On-Device Sentence Similarity for SMS Dataset}

\makeatletter
\newcommand{\linebreakand}{%
  \end{@IEEEauthorhalign}
  \hfill\mbox{}\par
  \mbox{}\hfill\begin{@IEEEauthorhalign}
}
\makeatother

\author{\IEEEauthorblockN{Arun D Prabhu }
\IEEEauthorblockA{\textit{On-Device AI} \\
\textit{Samsung R \& D Institute}\\
Bangalore, India \\
arun.prabhu@samsung.com}
\and
\IEEEauthorblockN{Nikhil Arora}
\IEEEauthorblockA{\textit{On-Device AI} \\
\textit{Samsung R \& D Institute}\\
Bangalore, India \\
n.arora@samsung.com}
\and
\IEEEauthorblockN{Shubham Vatsal}
\IEEEauthorblockA{\textit{On-Device AI} \\
\textit{Samsung R \& D Institute}\\
Bangalore, India \\
shubham.v30@samsung.com}
\linebreakand
\IEEEauthorblockN{Gopi Ramena}
\IEEEauthorblockA{\textit{On-Device AI} \\
\textit{Samsung R \& D Institute}\\
Bangalore, India \\
gopi.ramena@samsung.com}
\and
\IEEEauthorblockN{Sukumar Moharana}
\IEEEauthorblockA{\textit{On-Device AI} \\
\textit{Samsung R \& D Institute}\\
Bangalore, India \\
msukumar@samsung.com }
\and
\IEEEauthorblockN{Naresh Purre}
\IEEEauthorblockA{\textit{On-Device AI} \\
\textit{Samsung R \& D Institute}\\
Bangalore, India \\
naresh.purre@samsung.com }

}

\maketitle

\begin{abstract}
Determining the sentence similarity between Short Message Service (SMS) texts/sentences plays a significant role in mobile device industry. Gauging the similarity between SMS data is thus necessary for various applications like enhanced searching and navigation, clubbing together SMS of similar type when given a custom label or tag is provided by user irrespective of their sender etc. The problem faced with SMS data is its incomplete structure and grammatical inconsistencies. In this paper, we propose a unique pipeline for evaluating the text similarity between SMS texts. We use Part of Speech (POS) model for keyword extraction by taking advantage of the partial structure embedded in SMS texts and similarity comparisons are carried out using statistical methods. The proposed pipeline deals with major semantic variations across SMS data as well as makes it effective for its application on-device (mobile phone). To showcase the capabilities of our work, our pipeline has been designed with an inclination towards one of the possible applications of SMS text similarity discussed in one of the following sections but nonetheless guarantees scalability for other applications as well.
\end{abstract}

\begin{IEEEkeywords}
sentence similarity, word embeddings, keyword extraction, LSTM, deep learning, SMS, on-device 
\end{IEEEkeywords}

\section{Introduction}
Mobile devices today have become an inseparable part of our lives. With ever growing connectivity technology, mobile device ownerships are increasing rapidly. Thus a demand for device based Artificial Intelligent (AI) solutions is crucial in order to augment user experience. SMS texts are an important form of mobile device text data we come across in our day to day life. Ranging from various shopping offers to every sign in one time passwords (OTPs), SMS texts are in abundance. SMS being one of the most widely used mode for communication has an estimated 3.5 billion active users. But as SMS does not adhere to proper grammar rules and due to the unstructured nature of this data, not much work has been done in this domain. Any advancements done in this area would lay the foundation for many mobile device application based use cases and thus would impact a wide audience. Estimating the degree of similarity between these texts successfully would form the baseline for multiple Natural Language Processing (NLP) tasks in this domain.

We live in a world where the usage of hashtags is rampant on social media. These hashtags can be analysed in different manners serving as a useful feature for many NLP tasks, one of the most basic ones being classification of texts by treating a particular hashtag as one of the labels based on text similarity. We make use of a similar concept in our work. Cluttering of SMS inbox is a common scenario these days. It is very difficult to manually scroll and search for all kinds of transaction based OTPs corresponding to different vendors and view all of them at the same place. Similar to social media, in case of SMS also people can use hashtags to club together similar kind of texts or address privacy concerns associated with certain kind of SMS texts. In this paper we demonstrate a scenario where even minimal tagging of similar SMS data by users allow our proposed architecture to predict similar texts at a considerably impressive accuracy.

One of the questions which may arise after discussing about our use case could be Why are we not classifying SMS data into predetermined labels? Classification puts a restriction on the number of classes in which a given dataset can be divided, as can be seen in \cite{9031530}, where the classification of SMS data has been carried out with fixed number of classes. Every time a new label or class needs to be added, we need to retrain the entire model with the new class data. Another important factor to highlight in case of classification is that user intent is not captured and the classes are defined based on the observations of a handful of annotators. There can be a factor of subjectivity involved while considering two texts similar. For example, for a user $U_1$ SMS "ABC has added Rs.10 of cashback in your ABC wallet for your transaction at XYZ Foods. Updated balance: Rs.12." and "Dear Customer get Rs.10 off your first order and upto Rs.20 cashback with ABC. Use promo code FREEEAT to unlock. Details and TnC URL Order Now URL" are similar as they involve a common vendor ABC although the former message talks about a transaction in general whereas the latter message talks about an offer. Similarly for another user $U_2$, these two texts "Sweetest Treat from ABC's! Buy a Cheeseburst Pizza \& get a Choco Lava Cake at Just Rs10! Use Code 'FREEPIZZA'|Hurry! Walk-in /Order @App URL" and "Dear Customer get Rs.10 off your first order and upto Rs.20 cashback with XYZ Pvt Ltd Use promo code FREEOFF to unlock. Details and TnC URL Order Now URL" are similar as they both talk about offer although they contain different vendors ABC and XYZ Pvt Ltd. Our pipeline has been designed in a way to handle such user based subjectivities which can be further be to used to club similar data and in turn classify SMS data at a much granular level without the need of retraining any machine learning model explicitly again and again.

Many approaches have been designed for sentence similarity comparisons which make use of lexical as well as linguistic analysis. The objective of lexical matching mechanisms is to determine whether the given texts are alike in terms of lexical overlap, distance based similarity statistics etc. Linguistic analysis makes use of mechanisms like syntactic trees or dependency parsing. For handling text similarities, parsers are readily available, but not all parsers are qualitative or efficient enough. The challenge for most of the systems using text similarity is associated with scalability on mobile devices. The problem persists due to low Random Access Memory (RAM) and other resource constraints on such devices. 

In this paper, we propose a pipeline which combines insights from word level information to evaluate sentence similarity between two SMS. We use a lightweight POS \cite{lample2016neural,ma2016end} tagger to extract important keywords incorporated in the given SMS text based on its partial structural sense. Next, we use modified Bag of Words (BoW) %\cite{harris1954distributional} 
concept to account for word level semantics present in SMS data. Finally, we use a context based Ratcliff/ Obershelp algorithm \cite{ratcliff1988pattern} to account for sequence of words in the given texts. This pipeline apart from being capable of handling the grammatically inconsistent and incorrect structure also displays its potential in terms of low resource consumption for its deployment on mobile devices.

The remaining part of the paper is organized in the following
manner: Section II talks about the related works and how
our work differs from them; Section III describes the overall
proposed pipeline; Section IV provides the experiments conducted and their corresponding results; Section V talks about the other applications of this pipeline and Section VI finally talks about some of the improvements which can be incorporated in future.

\section{Related Work}

Predicting user-defined labels on incoming messages is an instance of incremental classification problem since the number of classes are not fixed. We aim to deploy such an application on-device where training of models is not possible. One trivial solution to this problem is to retrieve the most similar SMS in the labeled set and assign the label of the most similar SMS. Many previous works have explored short text similarity and have proposed various methods from statistical word based approaches to contextual similarity. Most of these papers focus on text in general or domain specific text. To the best of our knowledge, there is no prior work to compute similarity on SMS or short conversational messages. SMS data being improperly structured and often grammatically inconsistent, tend to difficult in any natural language task. 

Word vectors like Word2vec 
\cite{mikolov2013distributed}
, GloVe 
\cite{pennington2014glove} 
etc. capture word level semantics. These vectors can be used to compute similarity between words by using a similarity function like cosine. But, it is not evident how text sentences should be represented with them. Several approaches have been proposed for sentence-level and document-level semantics. Le and Mikolov \cite{le2014distributed} propose a variation on the Word2vec algorithm for computing paragraph vectors, by adding an explicit feature to the input of the neural network. SkipThought vectors \cite{kiros2015skip} propose an objective function that adapts the skip-gram model for words to the sentence level.

Many works require huge corpora for training and rely on external natural language resources. Recently, the state-of-the-art performance on naural language processing tasks are obtained by using deep contextualised word representations such as BERT \cite{devlin2018bert}, ELMo \cite{peters2018deep} or sentence encoders such as Infersent \cite{conneau2017supervised}, Universal Sentence Encoder \cite{cer2018universal}. These models either use transformer-based encoders %\cite{vaswani2017attention} 
or an attention mechanism on hidden states of LSTM \cite{hochreiter1997long}, 
%LSTM \cite{bahdanau2014neural}
making it computationally heavy and unfeasible to be deployed on-device. 

Other previous works use external knowledge from sources of structured semantic knowledge like Wikipedia and Wordnet \cite{miller1995wordnet} for text similarity tasks \cite{sedding2004wordnet, wang2008building}. To the best of our knowledge, none of the works focused on SMS data and on-device approaches. Due to device limitations, external knowledge from Wikipedia would be resource intensive. Computation of similarity using Wordnet is limited with its hierarchical representation. Word embeddings represents words in a multi-dimensional vector space enabling us to calculate the relationship between any two words in the corpus. In this paper, we present our work which promises good similarity results on SMS messages despite on-device limitations.

\begin{figure}
\centerline{\includegraphics[scale=0.22]{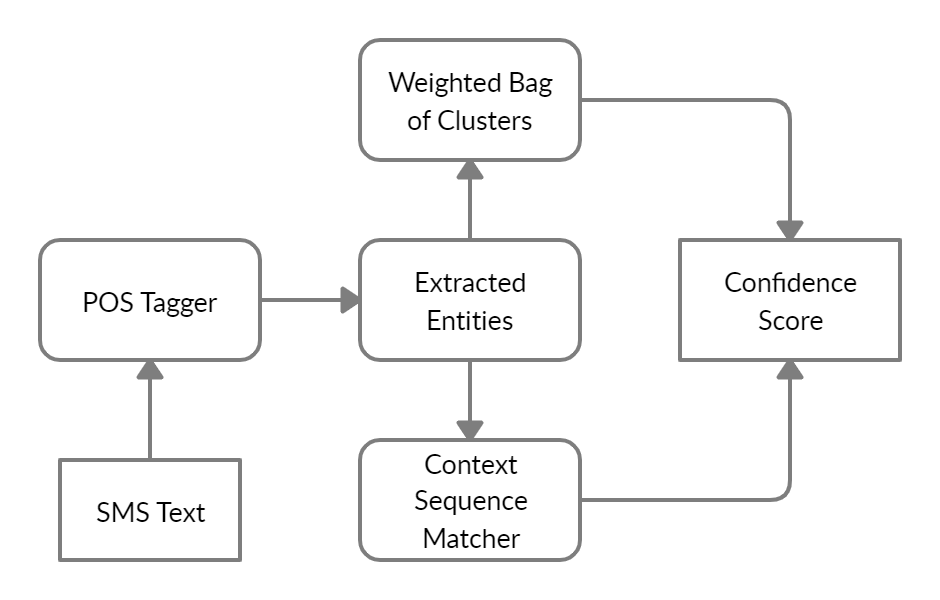}}
\caption{Proposed Pipeline}
\label{Flow}
\end{figure}

\section{Proposed Pipeline}
This section delineates the purpose of each component and eventually concludes how these components blend together to get us the desired result. Fig \ref{Flow} shows the outline of our method.

\subsection{Corpus Details}\label{sec: corpus}
We created our corpus by collecting around 7400 SMS texts which were then used for human annotation. The annotators judged the semantic and contextual similarity to segregate our data into multiple categories. The most prominent labels identified were Flight Travel, Debit Transaction,  Credit Transaction, Food Offer, Hotel Offer, Login OTP and Transaction OTP. The records were labelled by multiple annotators for considering the inter-annotator agreement. When we use these annotated records for our similarity and score analysis, each record is labelled in accordance with the class label which gets the maximum number of votes.
\begin{figure}
\centerline{\includegraphics[width=0.9\linewidth]{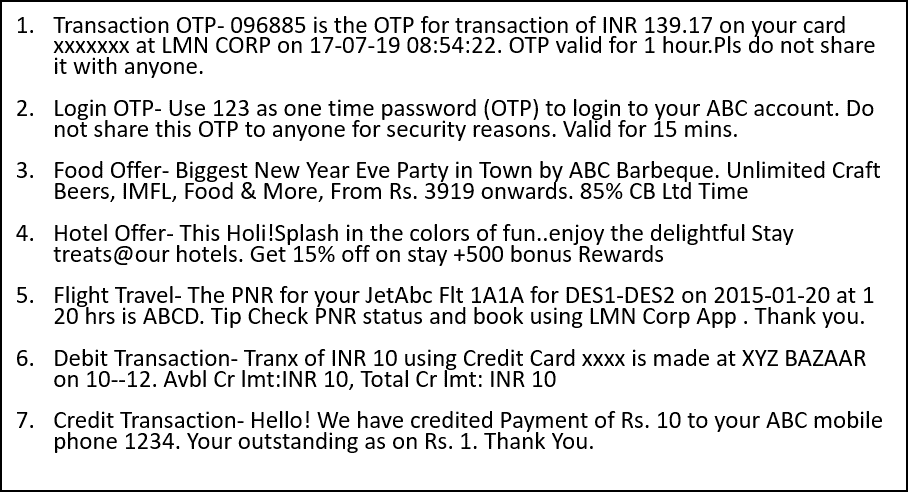}}
\caption{Data Snapshot}
\label{snap}
\end{figure}

In order to improve the annotation process, certain evaluation statistics were taken into account. We calculate the inter-annotator agreement for our set of records. One common mechanism for evaluating the inter-annotator agreement is the Kappa statistic \cite{mchugh2012interrater}.
%\cite{siegel1988case,ng1999case}
It is used in order to estimate the level of agreement between different human subjects. It also takes into account that the agreement between those subjects could be by coincidence. Suppose two human subjects are provided with ‘n’ identical samples for labelling, and they agree to a singular label on ‘a’ samples. So the level of agreement between the two of them would be
\begin{equation}
P_{a}\ =\ ( a/n)
\end{equation}
Now let us assume that there are ‘m’ distinguished class labels in our task and c\textsubscript{j} is the number of samples that both subjects identified as class 'j'. The probability that the two subject annotators agree by coincidence is evaluated by
\begin{equation}
P_{e}\ =\ \Sigma ^{m}_{j=1} \ ( c_{j/2} /N)^{2}
\end{equation}
The Kappa statistic $\displaystyle{\omega}$ is then defined as: 
\begin{equation}
\omega \ =\ ( P_{a}\ -\ P_{e}) /( 1\ -\ P_{e})
\end{equation}

We judged the level of agreement for pairwise annotator subjects using this method and the averaged result is shown in Table \ref{annotator_agreement}.  

\begin{table}[htbp]
\caption{The inter-annotator agreement analysis}
\begin{center}
\begin{tabular}{|M{5cm}|M{0.45cm}|M{0.45cm}|M{0.45cm}|}
\hline
%\textbf{Table}&\multicolumn{3}{|c|}{\textbf{Table Column Head}} \\
\textbf{Set of Labels} & \textbf{\textit{Pa}}& \textbf{\textit{Pe}}& \textbf{\textit{ $\displaystyle \omega$}} \\
\hline
Debit Transaction, Credit Transaction , Login OTP, Transaction OTP, Food Offer, Hotel Offer, Flight Travel   & 0.823& 0.244 & 0.765  \\
\hline
\end{tabular}
\label{annotator_agreement}
\end{center}
\end{table}

After following the above mentioned methodology for corpus collection, we ended up with around 300 SMS for each label. The snapshot of data for each label is shown in Fig. \ref{snap}

\subsection{Keyword Extraction}
\subsubsection{Part Of Speech (POS) model}

\begin{figure}
\centerline{\includegraphics[scale=0.6]{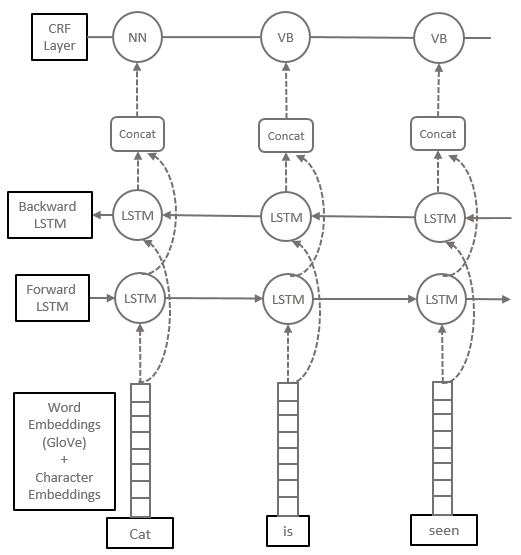}}
\caption{Part of Speech Model}
\label{PoS}
\end{figure}

For building a POS model, a model similar to Lample et al. \cite{lample2016neural} and Ma and Hovy \cite{ma2016end} is used. Firstly, a Bi-LSTM layer is trained to get character embeddings from the training data. This gives a character-based representation of each word. Next this is concatenated with standard GloVe (100 dimension vectors trained on 6 billion corpus of Wikipedia 2014 and Gigaword5) word vector representation. This gives us the contextual representation of each word. Then, a Bi-LSTM is run on each sentence represented by the above contextual representation. This final output from the model is decoded with a linear chain conditional random field using Viterbi algorithm. For on-device inference, the Viterbi decode algorithm is implemented in Java to be run on android devices and get the final output. Fig \ref{PoS} shows the architecture of our model.

We used the tagged dataset from the CoNLL-2003 shared task for training of the above neural network. The various parameters and metrics of the model are listed in Table \ref{POSMetrics}.

\begin{table}
\caption{Hyperparameters and metrics of POS model}
\begin{center}
\begin{tabular}{|c|c|c|c|}
\hline
\textbf{Hyperparameters} & \textbf{Value} \\
\hline
Word embeddings dimension & 100 \\
\hline
Char embeddings dimension & 50  \\
\hline
Num units in LSTM & 100  \\
\hline
Accuracy & 97.21\%  \\
\hline
\end{tabular}
\label{POSMetrics}
\end{center}
\end{table}

\subsubsection{Keyword identification}
We use POS model and extract nouns, proper nouns and verbs from the SMS. These parts of the SMS are the most essential since the other words are modifiers (adverbs and adjectives), connectors (conjunctions) or show relations (prepositions).  These keywords capture the essence of the SMS.

\subsection{Weighted Bag Of Clusters (WBoC)}
We have assumed that on an average a user tags 6 messages for each label. We create a WBoC using this minimal training set corresponding to each label. Each label consists of multiple clusters. Each cluster contains semantically similar words and is represented by two features, frequency and cluster embedding. The frequency is basically the number of occurences of words present in a cluster whereas the cluster embedding is the average word embeddings of all the words present in the cluster. We use GloVe (50 dimensions word vectors) to get the embeddings of words in a cluster. We optimally selected a threshold of 0.7 for cosine similarity amongst the words while forming a cluster. For out of vocabulary (OOV) words like vendor names, we form a singleton cluster with null embeddings. Fig \ref{BOW} gives a clearer picture of how bag of clusters are formed with respect to a label. We compute a weighted similarity score while measuring the degree of similarity between a test SMS $s_{1}$  and a user label $l_{1}$. The POS outputs of  $s_{1}$ can be denoted with $w_{i}$. Let $max(cos (c_{j}, w_{i}))$ represent the maximum cosine similarity value between the word $w_{i}$ of $s_{1}$ and the clusters $c_{j}$ of  $l_{1}$. If $f_{j}$ be the frequency of that corresponding cluster, then the weighted similarity score $WBoC(l_{1},s_{1})$ can be calculated by the equation

\begin{equation}
WBoC( l_{1}, s_{1}) = \frac{\sum_{i} ( f_{j} * max(cos (c_{j}, w_{i}) )) \ \forall \ c_j \in l_{1}}{\sum_{j} f_{j}}
\end{equation}

\subsection{Contextual Sequence Matcher (CSM)}
The final component of our pipeline is responsible to identify semantically similar localised sequences while computing sentence similarity between given texts. CSM makes use of a modified version of Ratcliff / Obershelp Pattern Recognition also known as Gestalt Pattern Matching \cite{ratcliff1988pattern}. According to the recognised version of Ratcliff/ Obershelp Pattern Recognition, the similarity $ Sim()$ between two sentences  $s_{1}$ and  $s_{2}$ is given by equation

\begin{equation}
Sim( s_{1},\ s_{2}) \ =\ 2* M( s_{1},  s_{2}) \ /\ ( |s_{1}| + |s_{2}|)
\end{equation}

where $M( s_{1},  s_{2})$ is the total number of matching characters and $ |s_{1}| $ and $ |s_{2}| $ are number of characters in $s_{1}$ and $s_{2}$ respectively. The matching characters are calculated as the longest common sub-string (LCS) plus recursively the number of matching characters in the non-matching regions on both sides of the LCS. We modified this definition of Ratcliff/ Obershelp Pattern Recognition in two ways. First we changed the existing definition to make it suitable for words instead of characters. Second, we included a factor of context using cosine similarity while checking the matching criteria for two words. So, according to the modified definition similarity $SimContx()$ between two sentences $s_{1}$ and $s_{2}$ can be written as

\begin{equation}
SimContx( s_{1} ,s_{2}) \ =\ 2*M_{w}( s_{1},  s_{2}) \ /\ ( |s_{1}| + |s_{2}|)|)
\end{equation}

where$M_{w}( s_{1},  s_{2})$ is the total number of matching words. The matching words are calculated using the same logic using LCS as explained above for characters. Let $s_{1}[w_{1},w_{2}..w_{n}]$ be a sentence of length n and  $s_{2}[w_{1},w_{2}..w_{m}]$ be a sentence of length m. Let $ s_{1}[i, r]$ or $s_{1}[w_{i}, w_{i+1}..w_{i+r}]$ represent a sequence of length r in $s_{1}$. Similarly, let $s_{2}[j,r$] or $s_{2}[w_{j},w_{j+1}..w_{j+r}]$ represent a sequence of length r in $s_{2}$. Also, let $ cos(x,y)$ represent cosine similarity value of words \emph{x} and \emph{y} using GloVe embeddings. Now $M_{w}( s_{1},  s_{2})$ can be written as

\begin{equation}
\label{Ratcliff}
\resizebox{1\hsize}{!}{$
M_{w}( s_{1}, s_{2})  = \begin{cases}
2*|seq| + L_{rec} + R_{rec} &  \begin{array}{l}
if\  s_{1}[ c] = s_{2}[ c] \ or \\
cos( s_{1}[ c] , s_{2}[ c])  \geqslant0.7\\
where\ 0 \leqslant c \leqslant  r
\end{array}\\
0 & \ otherwise
\end{cases}
$}
\end{equation}

%\begin{equation}
%M_{w}( s_{1}, s_{2})  = 
%\begin{cases}
%2 * |seq| + L_{rec} + R_{rec} & if \  \eqref{eq:cond} \\
%0 & otherwise
%\end{cases}
%\end{equation}

%where \eqref{eq:cond} represents the condition given by

%\begin{equation}
%\begin{split}
%if\  s_{1}[ c] = s_{2}[ c] \ or \ cos( s_{1}[ c] , s_{2}[ c])  \geq 0.7 \\
%where\ 0 \leq c \leq  r \label{eq:cond}
%\end{split}
%\end{equation}

%\begin{equation}
%\begin{split}
%s_{1}[ c] = s_{2}[ c] \ or \ cos( s_{1}[ c] , s_{2}[ c])  \geq 0.7
%\ where\ 0 \leq c \leq  r \label{eq:cond}
%\end{split}
%\end{equation}

where $ L_{rec} = M_{w}(s_{1}[0,i-1], s_{2}[0,j-1])$ and $R_{rec} = M_{w}(s_{1}[r+1,n], s_{2}[r+1,m])$. In the above equation $| seq|= | s_{1}[i,r]|=| s_{2}[j,r]|$. This component plays a significant role in identifying semantically similar sequences and identifying common OOV words which comprise mostly of vendor names. The condition $s_{1}[ c]=s_{2}[ c]$ checks if two same words are present which is especially helpful while checking same vendor names for which word embeddings are not present and hence cosine similarity cannot be calculated. Although our CSM does not take into consideration the exact positions of matching sequences, SMS dataset hardly observes contextual switches in the form of negations or other ways like we observe in case of a Sentiment Analysis. So, irrespective of the positions of sequences, the intent of any SMS text remains the same and the results as discussed in section (mention section name) showcases how this component is able to improve our results.

% \subsection{Weighted Bag of Clusters (WBoC)}
\begin{figure}
\centerline{\includegraphics[width=0.9\linewidth]{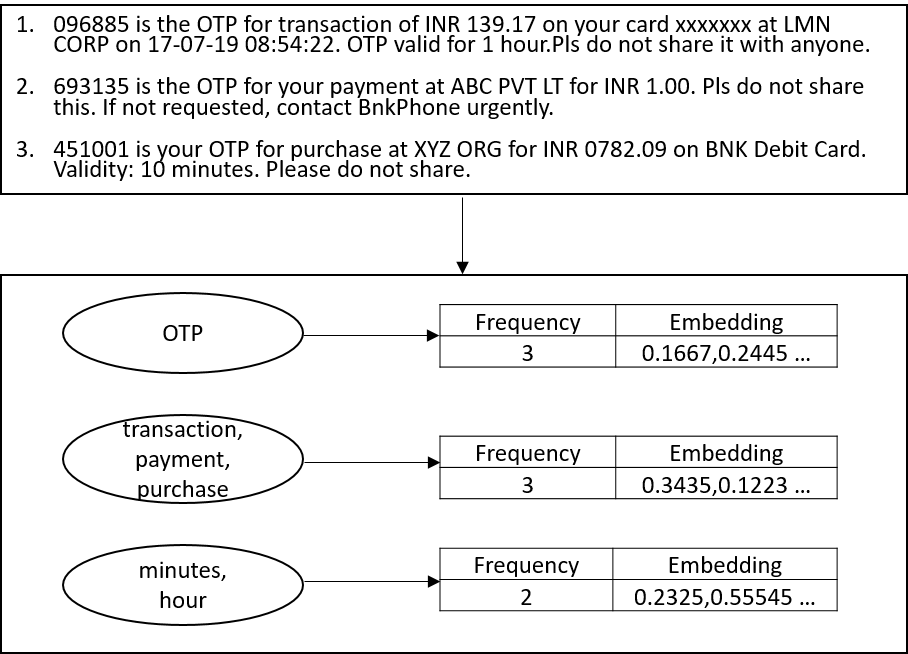}}
\caption{Working of WBoC}
\label{BOW}
\end{figure}

We compute the CSM score as given in equation \eqref{Ratcliff} between a given test SMS and all the user tagged SMS for a given label. The labelled SMS having highest similarity score with the given test SMS is considered as the final score of that test SMS with that label. We follow the same procedure across all the labels. Let \emph{s} represent tagged SMS belonging to label $l_1$. The final CSM score between a test SMS $s_1$ and label $l_1$ is given by 
\begin{equation}
CSM(s_{1}, l_{1}) \ =\ max( SimContx(s_{1}, \ s)) \ \forall \ s\ \in l_{1}\
\end{equation}

\subsection{Confidence Score}
The overall confidence of a sentence $s_1$ belonging to a particular label $l_1$ is calculated as:
\begin{equation}
\label{Total}
\begin{split}
\resizebox{.89\hsize}{!}{$
C( s_{1},l_{1})  = \alpha * WBoC(s_{1}, l_{1})\ + \ ( 1-\alpha )  * CSM(s_{1}, l_{1})$}
\end{split}
\end{equation}

where $\alpha$ is $<$ 1 and controls the relative effect that semantic similarity and word order values have on the overall confidence. \cite{wiemer2000adding} states that $\alpha$ values should be greater than 0.5 since word order plays a subordinate role in semantic processing of text. The label with the highest confidence is assigned to the sentence. 

\begin{table}
\caption{Performance of the model at different $\alpha$}
\begin{center}
\begin{tabular}{|c|c|c|c|c|}
\hline
\textbf{$\alpha$} & 0.6 & 0.7 & 0.8 & 0.9  \\
\hline
\textbf{Precision} & 0.4089	& 0.6171	& 0.8109	& 0.5601 \\
\hline
\textbf{Recall} & 0.2802    & 0.3645	& 0.6425    & 0.6904 \\
\hline
\textbf{F1 Score} &  0.2867	& 0.4138 & 0.6691 & 0.5954 \\
\hline
\textbf{Accuracy} & 0.4540	& 0.5867	& 0.7397	& 0.5816 \\
\hline
\end{tabular}
\label{alpha}
\end{center}
\end{table}

\section{Experiments \& Results}
\subsection{Evaluation}
We divided the data collected for each label in section \ref{sec: corpus} into two parts. First part consists of a small portion of data which is considered to be tagged by user. This portion of data is used as training set for WBoC. The other portion of data is used as test set. For each label, we considered around 2\% of data as user tagged whereas remaining 98\% is considered as test set. For evaluation purposes, we use a method similar to k-fold cross-validation. Since in our case the training data is less than test data, we modify k-fold cross validation to accommodate the changes. In a normal k-fold cross validation containing k groups of data, training is done on k-1 groups and testing is done on kth group. But in our case, we did the opposite by choosing to train on kth group and test on k-1 groups. The value of parameter  k is chosen as 60.

\subsection{ Comparison Results}
We compare our proposed pipeline with the following pipelines/architectures. We have considered the confidence score threshold as calculated by equation \ref{Total} to be 0.7 across all the architectures while taking a decision whether the given SMS belongs to a label or not. There can be many closely associated labels with subtle differences in their vocabulary. For example, lab
els like Transaction OTP and Login OTP share much of their vocabulary. This high degree of vocabulary overlap can lead to many false positives. In order to avoid such false positives, we chose a considerably high but balanced confidence threshold score. Also, we did not want to choose a very high confidence threshold score as that would have made similarity computation process very strict allowing only minor variations in test data with respect to training data. We conducted experiments with different $\alpha$ values and results are shown in Table \ref{alpha}. As we can see, lower $\alpha$ values  makes  the pipeline  more  rigid  and  decreases  the  generalisation  ability of  the  pipeline. Similarly, higher values of $\alpha$ makes the role of CSM fade away and hence again bringing the overall performance down. To obtain an appropriate balance, we chose $\alpha$ value as 0.8 for our pipeline and went ahead to compare with other architectures.

 \subsubsection{Baseline}
In this pipeline, we removed our CSM component to showcase the importance of capturing word order while computing sentence similarity. The pipeline only with POS and WBoC components is not able to capture the context based syntactic structure of sentence and the results in Table \ref{results} clearly show the substandard performance of this pipeline compared to other architectures. The $\alpha$ value as defined in equation \eqref{Total} is taken as 1 in order to nullify the effects of CSM.

\begin{table}
\caption{Comparison Results}
\begin{center}
\centering
\begin{tabular}{|M{1.23cm}|M{1cm}|M{1cm}|M{1.35cm} |M{1.35cm}|}
%\begin{tabular}{|c|c|c|c|c|}
\hline
%\textbf{Metrics} & \textbf{USE} & \vtop{\hbox{\strut \textbf{POS }} \hbox{\strut \textbf{+ WBOW}}} &\vtop{\hbox{\strut \textbf{POS + WBOW }} \hbox{\strut \textbf{+ Stasis}}} &\vtop{\hbox{\strut \textbf{POS + WBOW }} \hbox{\strut \textbf{+ CSM}}} \\
% \textbf{Metrics} & \textbf{USE} &\textbf{POS +  WBoC} & \textbf{POS + WBoC + Stasis} & \textbf{POS +  WBoC +  CSM} \\
\textbf{Metrics} & \textbf{USE} &\textbf{Baseline} & \textbf{ Stasis} & \textbf{Our Method} \\
\hline
Precision	& 0.6281	& 0.5653 	& 0.6481	& 0.8109  \\
\hline
 Recall	& 0.8073	& 0.7534  & 0.6587	& 0.6425	 \\
\hline
 F1 Score &	0.6708	& 0.5977 & 0.6354	& 0.6691	 \\
\hline
 Accuracy	& 0.6224  & 0.5408     	& 0.6530  &	0.7397  \\
\hline
\end{tabular}
\label{results}
\end{center}
\end{table}

 \subsubsection{Stasis}
In this pipeline, we compare our CSM to the word order similarity used in Stasis \cite{li2006sentence}. So, the pipeline now comprises of POS, WBoC and Stasis. To calculate the word order similarity, the sentences are converted into word vectors. A list of unique words is constructed and the position of these words in the sentence form the word vector. When a word does not appear in the sentence, the position of the word with the highest similarity to the missing word is used assuming it exceeds a pre-set threshold. Otherwise, the position is denoted by 0. We use GloVe embeddings to calculate the similarity between the 2 words. The threshold is set to 0.7. The value of $\alpha$ is set to 0.8 while calculating the similarity score for each test SMS. Table \ref{results} indicates that CSM in our pipeline performs better when it comes to capturing word orders and hence helps in achieving better results with respect to sentence similarity. We hypothesise that this is because Stasis uses stricter word order capturing mechanism which affects the generalising ability of the algorithm.

\begin{figure*}
\centering
\includegraphics[width=1\linewidth]{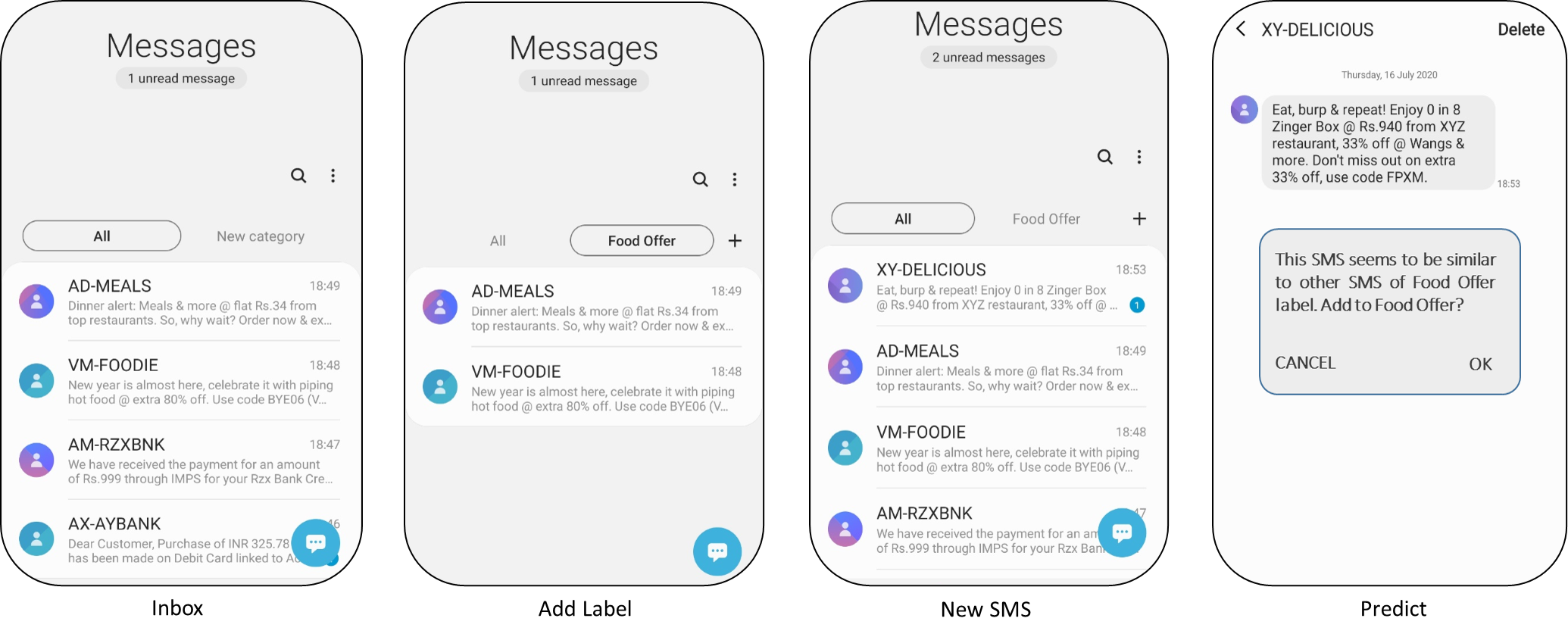}
\caption{On-Device Content Demonstration}
\label{fig:appflow}
\end{figure*}

\subsubsection{Sentence Encoder Architectures}
Recently, sentence encoders such as Infersent \cite{conneau2017supervised} and Universal Sentence Encoder\cite{cer2018universal} (USE) have come into picture which provide state of the art performance on the semantic textual similarity task. But the performance of the sentence encoders in the SMS domain is unknown. We evaluate Google’s USE lite on our SMS dataset for computing sentence similarity. USE outputs a 512-dimensional vector for any given sentence. We calculate cosine similarity between a given test SMS and all the tagged SMS of various labels. The user tagged SMS having highest similarity score with test SMS and crossing the similarity threshold  is assigned as the final label of tagged SMS. As we can see from the results in Table \ref{results}, even sentence encoders trained on millions of data are not able to appropriately capture the semantics of SMS data because of its grammatical inconsistencies and semi-structured nature. Another problem with sentence encoders are that most of their architectures are too computationally expensive for their on-device deployment and application. Since USE lite has been specifically designed for device based applications, we compared our pipeline results to USE lite.

\subsection{On-Device Metrics}
From the beginning we have emphasized on device related restrictions and hence the need to construct a corresponding solution. We have taken into consideration two commonly referred metrics to measure our pipeline's viability on-device. These metrics are inference time and pipeline size. The on-device metrics have been tabulated in Table \ref{ondevice} and have been calculated using Samsung’s Galaxy A51 with 4 GB RAM and 2.7 Ghz octa-core processor. As we can see, our pipeline consumes less than 30 MB size with inference time being as low as 281 ms and is still able to outperform USE.

\begin{table}
\caption{On-Device Metrics}
\begin{center}
\begin{tabular}{|c|c|c|}
\hline
\textbf{Component} & \textbf{Size} & \textbf{Inference Time (per Test SMS)} \\
\hline
POS & 7 MB & 60 ms \\
\hline
GloVe Embeddings &  18 MB & 18 ms \\
\hline
Complete Pipeline & 25 MB & 281 ms \\
\hline
\end{tabular}
\label{ondevice}
\end{center}
\end{table}

\section{Applications}
We have already discussed how our work revolves around one of the applications of predicting user labels. On-device screenshots illustrating the flow of this use case is demonstrated in Fig. \ref{fig:appflow}. Initially, user manually tags some SMS to a particular label. In our work, we have chosen a fixed quantity of 6 SMS as what is tagged by the user but for simplicity purposes, in the Fig. \ref{fig:appflow} it has been shown as only 2. Next when a similar kind of SMS arrives in SMS inbox and is read by the user, then user is informed about the similarity of this newly arrived SMS with that of SMS present under pre-defined label and is asked whether to move this SMS under the same label or not. This is one of the intuitive ways for content demonstration which focuses on improving user experience.

The text similarity work defined in this paper uses semantic as well as word order information and thus, can be used in various other on-device use cases as well. For example, in our work we have assumed user labels 6 SMS before we deploy our pipeline to predict test SMS. If we assume that we have only 1 user labelled SMS, then we can extend this architecture in searching and navigating across SMS inbox. 

%\subsection{Message Tagging with User Defined Tags}
%We take input messages and extract keywords from each one of them. Once the count reaches a certain number ‘X’, we will have sufficient records in our database, using which we can identify different classes for each message. We can provide the provision of defining the class types to the user itself, thus the classification will take user patterns into account. This is an efficient mechanism for on-device learning so we can use them to mark the messages as per our need.

%\subsection{Duplicate Message Detection}
%As we focus on text similarity, we can identify duplicate messages by identifying the context and clubbing those messages together. Thus, we will have a proper categorization of messages and the user can take action accordingly.

\section{Conclusion \& Future Work}
In this paper, we presented a unique pipeline to assign user defined labels to the new incoming messages. The important features of the proposed work can be summarized as
\begin{itemize}
    \item It identifies important words for comparing similarity using POS.
    \item It takes into consideration semantic word similarity by using word embeddings.
    \item It incorporates importance of word frequency while calculating weighted embeddings as a part of WBoC.
    \item It encompasses semantic word ordering using CSM.
    \item It is computationally efficient for on-device applications.
\end{itemize}

One important aspect that can be considered in order to improve our work is the user feedback. When the user disagrees with the labelling, we have to ensure that such messages are not labelled wrongly in the future. By collecting the feedback, we can easily integrate it in the components and provide better results with new learnings to the user.

\section{Acknowledgement}
The authors would like to thank all the users who contributed
in SMS data collection. The authors would
like to express their gratitude towards all the reviewers who
have given constructive feedback to improve the paper. In
addition, a special thanks to the annotators who followed a
rigorous set of rules in order to develop a consensus amongst
themselves while analysing the tags/labels for SMS dataset collected.

 \bibliographystyle{IEEEtran}
 \bibliography{IEEEabrv,references}

% \vspace{12pt}

\end{document}